\documentclass[]{acmart}
\def\BibTeX{{\rm B\kern-.05em{\sc i\kern-.025em b}\kern-.08emT\kern-.1667em\lower.7ex\hbox{E}\kern-.125emX}}

\usepackage{tikz}
\usepackage{collcell}

\usepackage{etoolbox}

\newtoggle{inTableHeader}
\toggletrue{inTableHeader}
\newcommand*{\StartTableHeader}{\global\toggletrue{inTableHeader}}%
\let\OldTabular\tabular%
\let\OldEndTabular\endtabular%
\renewenvironment{tabular}{\StartTableHeader\OldTabular}{\OldEndTabular\StartTableHeader}%

\newcommand*{\MinNumber}{0.0}%
\newcommand*{\MidNumber}{0.5} %
\newcommand*{\MaxNumber}{1.0}%

\newcommand{\ApplyGradient}[1]{%
  \iftoggle{inTableHeader}{#1}{
    \ifdim #1 pt > \MidNumber pt
        \pgfmathsetmacro{\PercentColor}{max(min(100.0*(#1 - \MidNumber)/(\MaxNumber-\MidNumber),100.0),0.00)} %
        \hspace{-0.33em}\colorbox{green!\PercentColor!yellow}{#1}
    \else
        \pgfmathsetmacro{\PercentColor}{max(min(100.0*(\MidNumber - #1)/(\MidNumber-\MinNumber),100.0),0.00)} %
        \hspace{-0.33em}\colorbox{red!\PercentColor!yellow}{#1}
    \fi
  }}

\newcolumntype{R}{>{\collectcell\ApplyGradient}c<{\endcollectcell}}

\setcopyright{none}

\newcommand\blfootnote[1]{%
  \begingroup
  \renewcommand\thefootnote{}\footnote{#1}%
  \addtocounter{footnote}{-1}%
  \endgroup
}

\begin{document}

\title{Task Preferences across Languages on Community Question Answering (CQA) Platforms}

\author{Sebastin Santy*}
\affiliation{
  \institution{Paul G. Allen School of Computer Science \& Engineering, University of Washington}
  \city{Seattle}
  \country{USA}}

\author{Prasanta Bhattacharya}
\affiliation{
  \institution{Institute of High Performance Computing (IHPC), A*STAR}
  \country{Singapore}
}

\author{Rishabh Mehrotra}
\affiliation{
 \institution{ShareChat}
 \city{London}
 \country{UK}}

 

\begin{abstract}

With the steady emergence of community question answering (CQA) platforms like Quora, StackExchange, and WikiHow, users now have an unprecedented access to information on various kind of queries and tasks. Moreover, the rapid proliferation and localization of these platforms spanning geographic and linguistic boundaries offer a unique opportunity to study the task requirements and preferences of users in different socio-linguistic groups. In this study, we implement an entity-embedding model trained on a large longitudinal dataset of multi-lingual and task-oriented question-answer pairs to uncover and quantify the (i) prevalence and distribution of various online tasks across linguistic communities, and (ii) emerging and receding trends in task popularity over time in these communities. Our results show that there exists substantial variance in task preference as well as popularity trends across linguistic communities on the platform. Findings from this study will help Q\&A platforms better curate and personalize content for non-English users, while also offering valuable insights to businesses looking to target non-English speaking communities online.
\end{abstract}


\maketitle

\section{Introduction}
\blfootnote{\color{red} \emph{Data collected and Analysis done in February 2020.}}
\blfootnote{*Work done while author was at University College London}
Users often seek specific information on the Internet to improve their performance on various tasks in their daily lives. For example, learning how to sing is a fairly complex task and hence, users often search online for various tutorials offering training or tips to help with specific aspects of singing. In recent years, several online platforms including how-to websites like wikiHow, Snapguide, eHow, Howcast have emerged as reliable sources of information to assist with solving these tasks. These websites act as how-to guides and detail the process of solving each task by deconstructing it into relatively simpler sub-tasks. Furthermore, solving one task often leads to creation of other related tasks, which are often serviced by articles recommended as follow-ups to the current article. Among such question answering (Q\&A) platforms, wikiHow has a particularly extensive presence online in terms of how-to guides across several domains and languages. 

In this study, we focus on characterizing and quantifying the task- and language-based heterogeneities prevalent in online Q\&A platforms, using wikiHow as an illustrative example. A recent report suggests that a little less than $60\%$ of the content available on Internet websites is in English\footnote{\url{https://w3techs.com/technologies/overview/content_language}}, even though only about $26\%$ of the internet users use English.\footnote{\url{http://www.internetworldstats.com/stats7.htm}} This has led to the creation of several language-localized communities including those on wikiHow that cater to the task-needs of non-English speaking users. As a consequence of the rapid expansion of these platforms, it has become imperative to translate these articles into destination languages, to increase their reach and popularity in non-English speaking countries. While such language-localized versions of platforms are becoming increasingly common (e.g. Wikipedia pages in non-English languages, wikiHow translations etc.), little is known about what kind of tasks are more or less likely to be translated into other popular languages, and the evolution of these translation activities over time.

\begin{figure}
\includegraphics[width=0.49\linewidth]{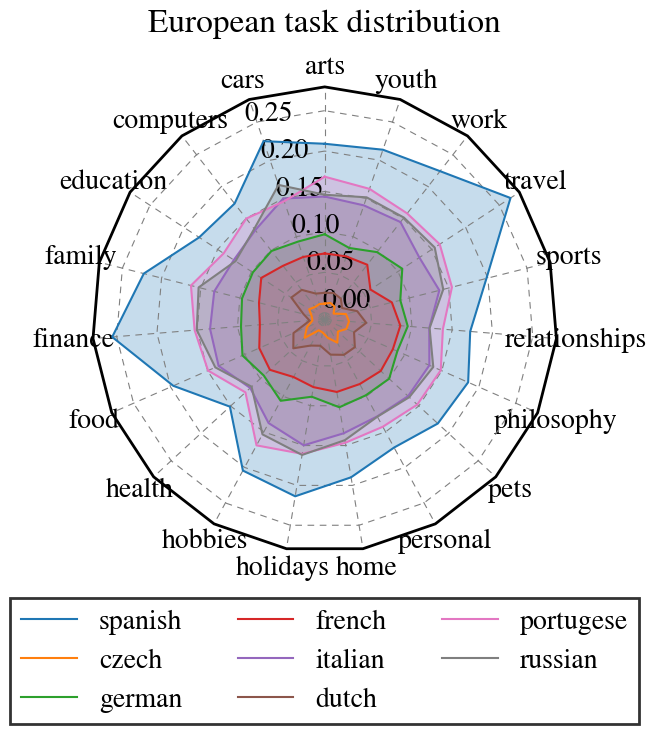}\hfill
\includegraphics[width=0.49\linewidth]{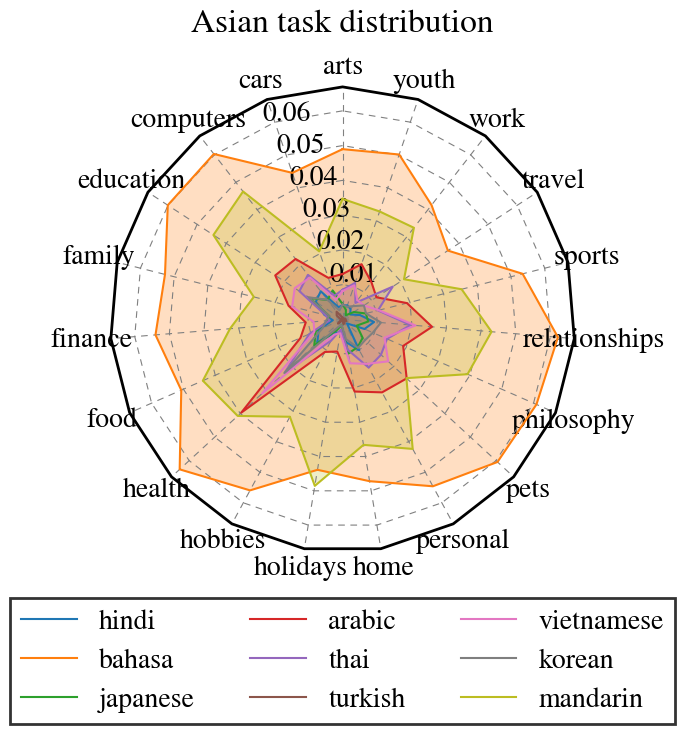}
\caption{Task-Language Distribution}
\label{fig:prelim}
\end{figure}
The characterization of user tasks has been a popular area of study in information retrieval, with recent work focusing on task understanding, extraction, ranking and prediction in various emerging contexts \cite{rishabh16, mehrotra2017extracting}. However, what remains to be examined is how these user tasks are distributed across various communities on digital platforms as well as the emerging trends in these task distributions. For community-based question answering (CQA) platforms, like wikiHow, we expect to see significant variation in the distribution of topics, tasks and task-categories across different languages represented on the platform. Moreover, since the topics on these platform directly correlate to actual tasks in the real-world, investigating such variances in task-representation across languages offers a unique opportunity to understand task popularity among users in various regions of the world. Such analysis will help platform owners to better forecast user preferences and behavior on their platform, which, in turn, will enable them to plan and provision their content curation activities and associated resources more effectively. Furthermore, the findings from this study also offer a unique window for computational social scientists to study the penetration and popularity of various categories of human activities in language-communities across the world. To our knowledge, this is among the first efforts at characterizing the distribution of task-related content across language communities on the Internet.

In this paper, we offer a characterization of the various task categories that are represented in articles spanning 17 different languages on the wikiHow platform. We use a recent entity embedding framework to generate two quantifiable metrics -- \textit{language inclusivity} and \textit{task preference}, which capture the heterogeneities in task translation across language communities by focusing on the distribution of tasks within particular language communities, and vice-versa. Using these metrics, we are able to uncover insights on how certain task categories like \textbf{computers}, \textbf{health}, and \textbf{education} (but not others) diffuse rapidly into other language communities over time. Moreover, we observe that certain task categories like \textbf{arts} and \textbf{finance} show greater prevalence in specific language communities (e.g. Spanish and Portuguese), than others.  

In the following section, we review related work on modeling search queries, sessions and tasks on digital platforms. Following this, we introduce our empirical context as well as our associated modeling strategy. Next, we present the results from our model estimation, followed by a brief discussion of the key findings. Finally, we conclude with a summary of limitations, and an overview of our future research plans.

\section{Related Work}
Prior research have focused on studying online search "sessions" when uncovering insights about user behavior. More recently, search "tasks" have emerged as an alternate source of such information. Specifically, recent studies have attempted to model in-session tasks \cite{jones2008beyond,lucchese2011identifying}, and cross-session tasks \cite{wang2013learning,kotov2011modeling} from query sequences using classification and clustering based approaches. Other studies have addressed the problem of cross-session task extraction using binary same-task classification \cite{kotov2011modeling, agichtein2012search}. More recent work has proposed methods for extracting task-subtask clusters~\cite{rishabh16}and task hierarchies \cite{mehrotra2017extracting}. These extracted tasks can consequently be used to predict task level user satisfaction \cite{mehrotra2017deep}. Beyond search queries, task information is now being used to encode how-to knowledge about problem solving tasks, by leveraging content from online
communities such as wikiHow~\cite{chu2017distilling}. Our current work extends this ongoing stream of research on tasks by uncovering the interplay between tasks and languages, in order to understand how tasks in community Q\&A sites evolve across language communities.

\section{Data and Context}
To uncover and analyze the distribution of tasks prevalent in various linguistic communities on the Internet, we sourced data from wikiHow, which is an online how-to guide for solving simple to complex tasks in a structured manner. Recent studies have emphasized the utility of wikiHow and other how-to communities in solving real-world tasks \cite{chu2017distilling}. 
wikiHow is an open-source and open-content community, where wiki-style articles are translated manually to any of the listed languages. Interestingly, these translations are done at a sentence-level, rather than at an article level, as with Wikipedia. We assume that the volume and frequency of translations for an article to a target language will correlate with the level of interest for the article in that language community.

To curate our dataset, we included all articles available on wikiHow \footnote{https://www.wikihow.com/} along with the different information stubs available on an article page, like the viewer statistics, languages it has been translated to, article content etc. We gathered nearly 138,000 articles, each of which represents the solution to a specific real-world task. These articles are categorized as a hierarchy of around 8000 nodes with a maximum depth of 8. In order to collect temporal data, we used the Wayback Machine.\footnote{https://archive.org/web/} The last snapshots from each year are recorded for each article.

The extracted tasks span the following categories: Arts and Entertainment (\textbf{arts}), Cars and Vehicles (\textbf{cars}), Computers and Electronics (\textbf{computers}), Education and Communications (\textbf{education}), Family Life (\textbf{family}), Finance and Business (\textbf{finance}), Food and Entertainment (\textbf{food}), Health (\textbf{health}), Hobbies and Crafts (\textbf{hobbies}), Holidays and Traditions (\textbf{holidays}), Home and Garden (\textbf{home}), Personal Care and Style (\textbf{personal}), Pets and Animals (\textbf{pets}), Philosophy and Religion (\textbf{philosophy}), Relationships (\textbf{relations}), Sports and Fitness (\textbf{sports}), Travel (\textbf{travel}), Work World (\textbf{work}), Youth (\textbf{youth}).

The 17 non-English languages included are Portugese (\textbf{por}), Spanish (\textbf{spa}), French (\textbf{fre}), Dutch (\textbf{dut}), German (\textbf{ger}), Italian (\textbf{ita}), Czech (\textbf{cze}), Russian (\textbf{rus}), Turkish (\textbf{tur}), Arabic (\textbf{ara}), Hindi (\textbf{hin}), Mandarin (\textbf{man}), Thai (\textbf{tha}), Vietnamese (\textbf{vie}), Bahasa (\textbf{ind}), Korean (\textbf{kor}), Japanese (\textbf{jpn}).

\section{Model}
\begin{figure}
    \centering
    \includegraphics[width=0.5\textwidth]{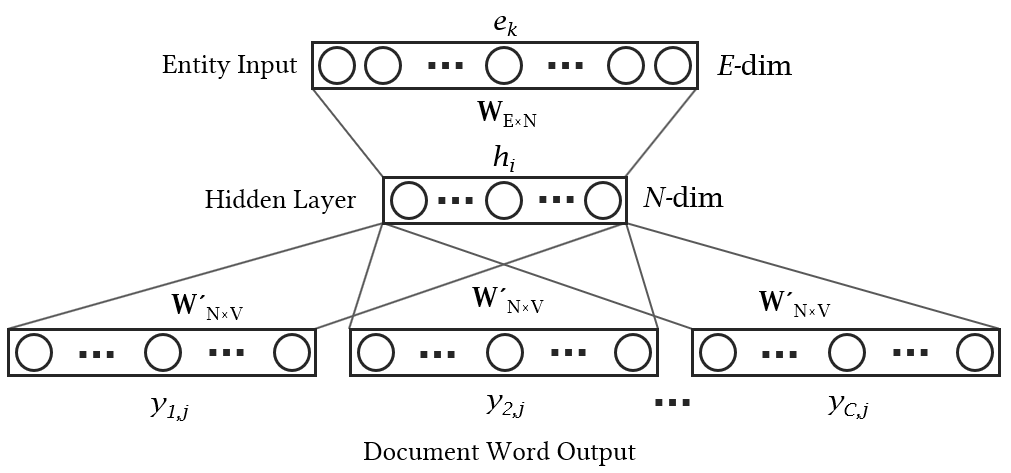}
    \caption{Entity Embeddings Model Architecture}
    \label{fig:model}
\end{figure}
Before we introduce our modeling strategy, we present some model-free analysis to better illustrate our data context. In order to understand the interplay between language and tasks over multiple years, we use 3 primary attributes of data: Time, Task Category, Languages. Figure \ref{fig:prelim} shows the distribution of languages over different task categories - normalized by the number of articles present in each task category. European languages are consistently associated with all task categories, whereas the Asian languages show clear preferences for certain task categories over others.



\begin{figure*}
    \centering
    \includegraphics[width=\textwidth]{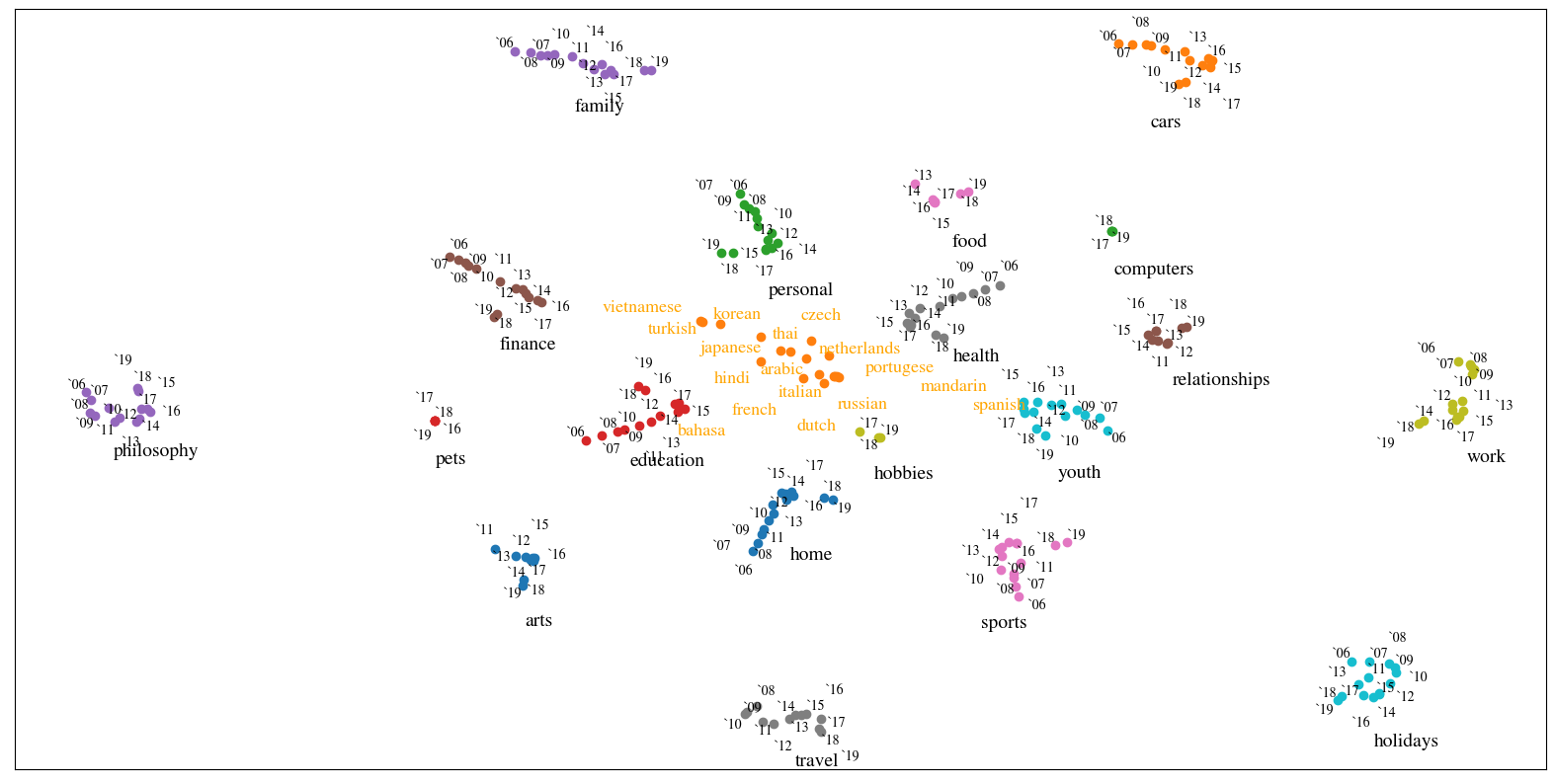}
    \caption{Task-Language Embeddings t-SNE Visualization: \small{\normalfont
    Each point in this visualization represents an entity which is learnt using the entity embeddings approach. It can be observed that the languages have clustered themselves in the center, and have arranged themselves according to their geolocation where they are spoken, thus suggesting a task variance among these communities. The shift of these task categories through different years is also apparent from the long trails which are pointing towards the center.}}
    
    \label{fig:embeddings}
\end{figure*}
In order to capture all three of the above-mentioned attributes together, we use an entity embeddings approach. Embeddings are known to capture inherent signals which frequency-based analyses often fail to reflect. Moreover, with entity embeddings we are able to build efficient representations which can help in understanding pattern continuity, if any, as well as drive prediction of how these tasks might be evolving.


Document to vector models~\cite{le2014distributed} have shown how documents can be represented in one space by modeling the relationship between texts. We follow a similar approach in order to build these entity embeddings~\cite{joshi-etal-2020-state}. Here, each document is an article from wikiHow across different tasks and years, as described in the data section. We use each task in a particular year (task\_year) and the translated languages as entities which represent each article.

We use a model which is similar to a Skip-Gram model with one hidden layer as shown in Figure \ref{fig:model}. The input layer accepts the entities and has size equal to the total number of entities ($E$-dim) (languages + task\_year). The hidden layer is the size of the embedding desired ($N$-dim), which we appropriately tune on a held-out data. The output layer predicts the words for the input entity and is of the same size of the vocabulary built from document texts. $W_{E\times N}$ is the weight matrix for the entity layer to the hidden layer. Similarly, $W_{N\times V}$ is the weight matrix for the hidden layer to the output layer computation. At the end of the training, $W_{E\times N}$ is the matrix containing entity embeddings and $W_{N\times V}$ is the matrix containing the word embeddings. To train the model, a single entity of an article is given as input, and the model is asked to predict the words from the contents of the article. The collective representation will help us understand how tasks have varied over the years along with the trends in translation to different tasks. We can also generate the embeddings of the words in the same space, which we can subsequently plot to visualize the relations.


\section{Analysis}

\begin{figure*}
    \centering
    \includegraphics[width=\textwidth]{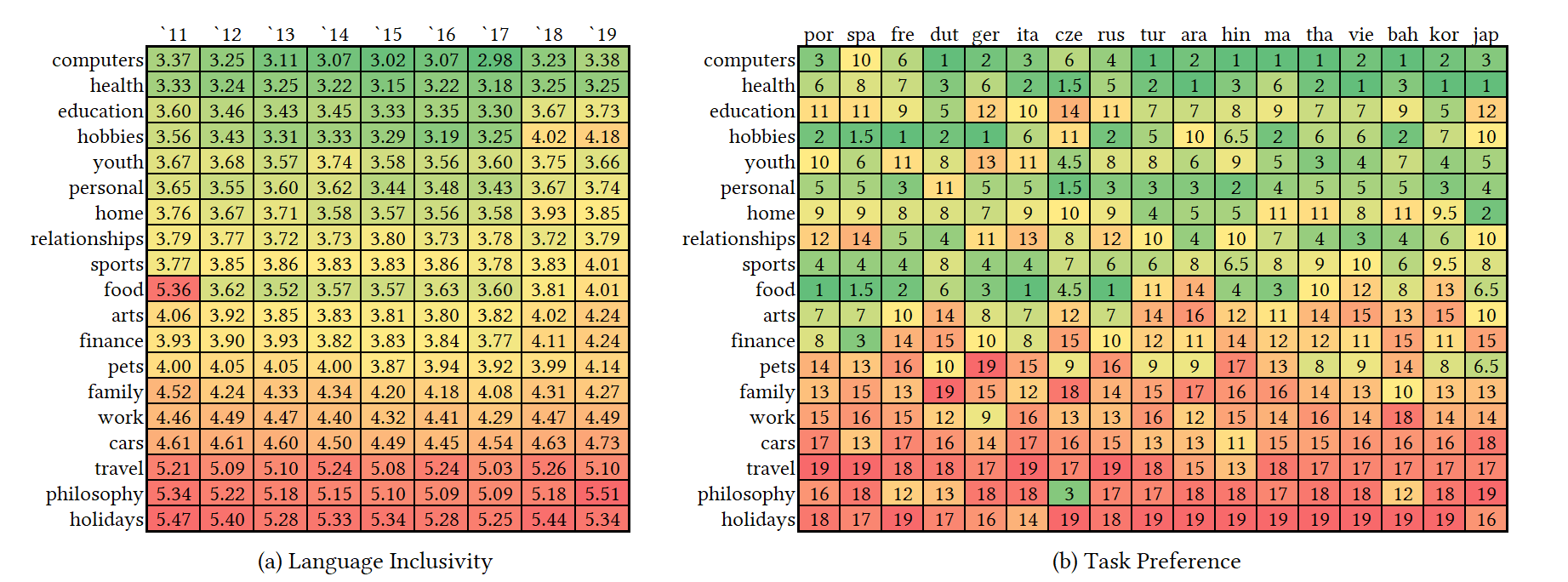}
    \caption{Metrics Computed using Entity Embeddings: \small{\normalfont
    (a) Language Inclusivity is the cosine distance between the average language vector and task of specific years. (b) Task Preference is the rank of cosine distance between the average task vectors and languages respectively.}}
    \label{fig:embeddings_metrics}
\end{figure*}

The model learns entity and document embeddings in the same space and of the same dimensions. For the current setup, we use 100 dimensions, and visualize it in 2 dimensions using t-SNE \cite{maaten2008visualizing}. Figure \ref{fig:embeddings} shows the entity embeddings represented in the same space. There are a number of interesting observations which can be made from just a visual inspection. However, model metrics can help to better quantify these insights.

\subsection{Language Inclusivity}
Characterizing the co-evolution of tasks and languages over the years will help us gain valuable insights. For instance, we can uncover tasks that are language-inclusive, and hence, benefit communities globally. It is clear from Figure \ref{fig:embeddings} that certain task categories show stronger acceleration towards the languages, as illustrated by their long trails. Conversely, certain other tasks have just formed a dense cluster, and barely exhibit any movement. We quantify this \textit{Language Inclusivity} by measuring cosine distance between the average of all language embeddings and tasks in specific years. Cosine distance is preferred over Euclidean distance in order to avoid frequency bias, thereby effectively helping to normalize the occurrences. Figure \ref{fig:embeddings_metrics} (a) shows these cosine distances. We can observe how certain task categories (e.g. \textbf{computers}) have a consistently high propensity of getting translated to multiple languages over the years, while other task categories (e.g. \textbf{travel}) exhibit the reverse. This helps us understand the spectrum of tasks where localization of content would matter. One of the key takeaways is that the task categories which are positively evolving towards language inclusion over the years, are also often categories which relate to the personal growth of individuals.

\subsection{Task Preference}
In the previous analysis, we observed how tasks are increasingly adapting to new languages. It would be equally interesting to observe, from the perspective of a community, how certain tasks are important to them. In order to quantify this, we calculate \textit{Task Preference} as the cosine distance between languages and task embeddings (\textit{which are averaged over the years}). For each language, we rank these cosine distances to understand the tasks which the community is particularly interested in.

The fair correlation between Figure \ref{fig:embeddings_metrics} (a) and (b) shows that, for most parts, the growth/expansion of tasks in different languages is organic. However, we can observe certain characteristics that hint at the presence of localized interests in specific communities which might be driving the translation of articles in those categories. For example - \textbf{food} shows a modest trend across the years in (a). However, in (b) it can be noticed as a very popular topic among the European language communities, but less popular among their Asian counterparts. Similarly, \textbf{computers} is a hot topic category among the Asian Languages ranging from Turkish to Japanese, but is less popular among their European counterparts.

\section{Conclusion}
This preliminary study is among the first to analyze inherent task- and language-level heterogeneities in online CQA platforms. We implemented an entity embedding model using a longitudinal dataset on wikiHow articles to uncover and quantify these heterogeneities prevalent on the platform. We also noted inherent differences in task distributions across language communities. Finally, our analyses also revealed certain unusual and counter-intuitive findings that warrant further investigation. For example, the \textbf{food} category shows substantial variance in its translation propensity over the years. In terms of prevalence, we noticed a strong variance across language communities. This is an interesting area for future investigation, as the quest for food-related tasks tends to be a common theme of interest for users worldwide. In future work, we will also look at specific aspects within each task, to understand what type of activities within each task category are prioritized in the translation process. Lastly, it might also be useful to benchmark some of our findings against other popular CQA platforms like Stack Exchange and Quora, to generate prescriptive insights for respective platform owners, as well as for businesses that benefit from user engagement on these platforms.


\bibliographystyle{ACM-Reference-Format}
\bibliography{sample-base}

\end{document}